\documentclass[10pt,twocolumn,letterpaper]{article}
\pdfoutput=1

\usepackage{iccv}
\usepackage{times}
\usepackage{epsfig}
\usepackage{graphicx}
\usepackage{amsmath}
\usepackage{amssymb}

\usepackage{array}
\newcolumntype{H}{>{\setbox0=\hbox\bgroup}c<{\egroup}@{}}
\usepackage{ftnright}
\usepackage{outlines}
\usepackage{pifont}
\newcommand{\cmark}{\ding{51}}%
\newcommand{\xmark}{\ding{55}}%
\usepackage{caption}
\usepackage{subcaption}
\usepackage{algorithm}
\usepackage{algorithmicx}
\usepackage{algpseudocode}
\usepackage{listings}
\makeatletter
\@namedef{ver@everyshi.sty}{}
\makeatother
\usepackage[normalem]{ulem}

\usepackage{color}
\usepackage{colortbl}
\usepackage{amsfonts}

\usepackage[accsupp]{axessibility}

\usepackage{dblfloatfix}
\usepackage[utf8]{inputenc}
\makeatletter
\@namedef{ver@everyshi.sty}{}
\makeatother
\usepackage{pgfplots}
\DeclareUnicodeCharacter{2212}{−}
\usepgfplotslibrary{groupplots,dateplot}
\usetikzlibrary{patterns,shapes.arrows}
\pgfplotsset{compat=newest}

\usepackage[breaklinks=true,bookmarks=false]{hyperref}

\usepgfplotslibrary{colorbrewer}

\iccvfinalcopy 

\def\iccvPaperID{6598} 

\ificcvfinal\fi

\newcommand{\ta}{TA}
\newcommand{\trivialaugment}{TrivialAugment}

\begin{document}


\title{TrivialAugment: Tuning-free Yet State-of-the-Art Data Augmentation}

\author{Samuel G. M\"uller\\
University of Freiburg\\
{\tt\small muellesa@cs.uni-freiburg.de}
\and
Frank Hutter\\
University of Freiburg \& \\
Bosch Center for Artificial Intelligence, Germany\\
{\tt\small fh@cs.uni-freiburg.de}
}

\maketitle
\ificcvfinal\thispagestyle{empty}\fi

\begin{abstract}
Automatic augmentation methods have recently become a crucial pillar for strong model performance in vision tasks.
While existing automatic augmentation methods need to trade off simplicity, cost and performance, we present a most simple baseline, \trivialaugment{}, that outperforms previous methods for almost free.
\trivialaugment{} is parameter-free and only applies a single augmentation to each image.
Thus, \trivialaugment{}\textquotesingle{s} effectiveness is very unexpected to us and
we performed very thorough experiments to study its performance.
First, we compare \trivialaugment{} to previous state-of-the-art methods in a variety of image classification scenarios.
Then, we perform multiple ablation studies with different augmentation spaces, augmentation methods and setups
to understand the crucial requirements for its performance.
Additionally, we provide a simple interface to facilitate the widespread adoption of automatic augmentation methods, as well as our full code base for reproducibility\footnote{\url{https://github.com/automl/trivialaugment}}.
Since our work reveals a stagnation in many parts of automatic augmentation research,
we end with a short proposal of best practices for sustained future progress in automatic augmentation methods.
\end{abstract}

\begin{figure}[!h]
\vspace{-.6cm}
    \centering
    \input{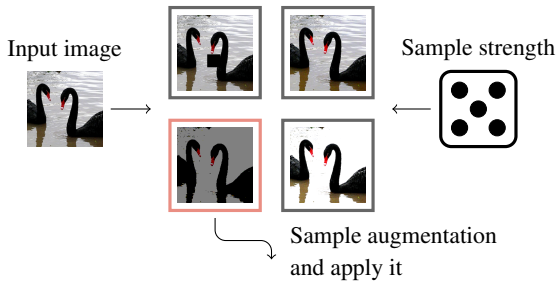}
    \caption{A visualization of \ta{}. For each image, \ta{} (uniformly) samples an augmentation strength and an augmentation. This augmentation is then applied to the image with the sampled strength.}
    \label{fig:tavisualisation}
\vspace{-.6cm}
\end{figure}

\section{Introduction}
\label{section:introduction}

Data Augmentation is a very popular approach to increase generalization of machine learning models by generating additional data.
It is applied in many areas, such as machine translation \cite{fadaee2017dataaugformt}, object detection \cite{Detectron2018} or semi-supervised learning \cite{UDAdataaugmentationsemisupervised}.
In this work, we focus on the application of data augmentation to image classification \cite{devries2017improved,krizhevsky-nips12}.


\begin{table}[!t]
    \setlength{\tabcolsep}{2pt}
    \small
    \centering
    \begin{tabular}{l|c|cccc}
    \vspace{0.00cm}
    \hspace*{-0.1cm}
         Method & Search & CIFAR-10 & CIFAR-100 & SVHN & ImageNet \\
                & Overhead & ShakeShake & WRN & WRN & ResNet \\
        \hline
         AA & $40$ - $800\times$ & 98.0 & 82.9 & 98.9 & 77.6\\
         RA & $4$ - $80\times$ & 98.0 & 83.3 & \textbf{99.0} & 77.6\\
         Fast AA & $1\times$ & 98.0 & 82.7 & 98.8 & 77.6\\
         \hline
         TA (ours) & {$\boldsymbol{0\times}$} & \textbf{98.2} & \textbf{84.3} & 98.9 & \textbf{78.1}\\
    \end{tabular}
    \caption{\textbf{\trivialaugment{} compares very favourably to previous augmentation methods.} In this table we summarize some results from Table \ref{tab:aug_results} and present augmentation search overhead estimates.}
    \label{tab:summary_table}
\end{table}

Image augmentations for image classification generate novel images based on images in a dataset, which are likely to still belong to the same classification category.
This way the dataset can grow based on the biases that come with the augmentations.
While data augmentations can yield considerable performance improvements, they do require domain knowledge.
An example of an augmentation, with a likely class-preserving behaviour, is the rotation of an image by some small number of degrees.
The image's class is still recognized by humans and so this allows the model to generalize in a way humans expect it to generalize.

Automatic augmentation methods are a set of methods that design augmentation policies automatically.
They have been shown to improve model performance significantly across tasks \cite{cubuk2020randaugment,zoph2019objectdetectionaugment,UDAdataaugmentationsemisupervised}.

Automatic augmentation methods have flourished especially for image classification in recent years \cite{cubuk2020randaugment,sungbin2019fastautoaugument,Lin2019ohlautoaug,ho2019populationbasedaugmentation} with many different approaches that learn policies over augmentation combinations.
The promise of this field is to learn custom augmentation policies that are strong for a particular model and dataset.
While the application of an augmentation policy found automatically is cheap, the search for it can be much more expensive than the training itself.

In this work, we challenge the belief that the resulting augmentation policies of current automatic augmentation methods are actually particularly well fit to the model and dataset.
We do this by introducing a trivial baseline method that performs comparably to more expensive augmentation methods without learning a specific augmentation policy per task.
Our method does not even combine augmentations in any way.
We fittingly call it \trivialaugment{} (\ta{}).

The contributions of this paper are threefold:
\begin{itemize}
    \item We analyze the minimal requirements for well-performing automatic augmentation methods and propose \trivialaugment{} (\ta{}), a trivial augmentation baseline that poses state-of-the-art performance in most setups. At the same time, \ta{} is the most practical automatic augmentation method to date.
    \item We comprehensively analyze the performance of \ta{} and multiple other automatic augmentation methods in many setups, using a unified open-source codebase to compare apples to apples.
    \item We make recommendations on the practical usage of automatic augmentation methods and collect best practices for automatic augmentation research.
    Additionally, we provide our code for easy application and future research.
\end{itemize}




\section{Related Work} 
Many automatic augmentation methods have been proposed in recent years with multiple different setups.
Still, all automatic augmentation methods we consider share one property:
They work on augmentation spaces that consist of i) a set of prespecified augmentations $\mathcal{A}$ and ii) a set of possible strength settings with which augmentations in $\mathcal{A}$ can be called (in this work $\{0,\dots,30\}$).
One member of $\mathcal{A}$ might, for example, be the aforementioned rotation operation, where the strength would correspond to the number of degrees.
Automatic augmentation methods now learn how to use these augmentations together on training data to yield a well-performing final classifier.

In this section, we provide a thorough overview of relevant previous methods.
As the compute requirements for automatic augmentation methods can dominate the training costs, we order this recount by the total cost of each method.

We begin with the first automatic augmentation method, AutoAugment (AA)~\cite{cubuk2019autoaugment}, which also happens to be the most expensive, spending over half a GPU-year of compute to yield a classifier on CIFAR-10.
AA uses a recurrent neural network (RNN), which is trained with reinforcement learning methods, to predict a parameterization of augmentation policies.
Reward is given for the validation accuracy of a particular model trained on a particular dataset with the predicted policy.
AA makes use of multiple sub-policies each consisting of multiple augmentations, which in turn are applied sequentially to an input image. 
Additionally, augmentations are left out with a specified probability. This allows one sub-policy to represent multiple combinations of augmentations.
Since AA is costly it uses not the task at hand for augmentation search, but a reduced dataset and a smaller model variant.

The second most expensive method is Augmentation-wise Sharing for AutoAugment (AWS) \cite{tian2020awsimprovingautoaug}.
It builds on the same optimization procedure as AA, but uses a simpler search space.
The search space consists of a distribution over pairs of augmentations that are applied together.
Different from AA, AWS learns the augmentation policy for the last few epochs of training only.
It does this on the full dataset with a small model.

A very different approach, called Population-based Augmentation (PBA)~\cite{ho2019populationbasedaugmentation}, is to learn the augmentation policy online as the training goes.
PBA does so by using multiple workers that each use a different policy and are updated in an evolutionary fashion.
It uses yet another policy parameterization: a vector of augmentations where each augmentation has an attached strength and leave-out probability.
From this vector augmentations are sampled uniformly at random and applied with the given strength or left out, depending on the leave-out probability.

Another method based on multiple parallel workers is Online Hyper-Parameter Learning for Auto-Augmentation (OHL) \cite{Lin2019ohlautoaug}.
Here, the policy is defined like for AWS and its parameters are trained using reinforcement learning.
The major difference with AWS is that its reward is the accuracy on held-out data after a part of training like for PBA, rather than final accuracy.
As an additional way of tuning the neural network weights in the parallel run, the weights of the worker with maximal accuracy are used to initialize all workers in the next part of training.

\newcommand{\advaa}{Adv.\ AA}
Adversarial AutoAugment (\advaa{}) \cite{zhang2020adversarialaa} is another slightly cheaper method that uses multiple workers and learns the augmentation policy online.
It trains only a single model, though.
Here, a single batch is copied to eight different workers and each worker applies its own policies to it, similar to the work by Hoffer \etal \cite{Hoffer20_AugmentYourBatch}.
The worker policies are sampled at the beginning of each epoch from a policy distribution.
The policy distribution has a similar form to that of AA.
After each epoch, \advaa{} makes a reinforcement-learning based update and rewards the policy yielding the \textit{lowest} accuracy training accuracy, causing the policy distribution to shift towards progressively stronger augmentations over the course of training.

Recently, Cubuk \etal proposed RandAugment (RA) \cite{cubuk2020randaugment}. It is much simpler, but only slightly cheaper, compared to the previous methods. 
RA only tunes two scalar parameters for each task:
(i) a single augmentation strength $m \in \{0,\dots,30\}$ which is applied to all augmentations and
(ii) the number of augmentations to combine for each image $n \in \{1,2,3\}$.
RA therefore reduces the number of hyper-parameters from all the weights of an RNN (for AA) or a distribution over more than a thousand augmentation combinations (for AWS and OHL) to just two.
This radical simplification, contrary to expectations, does not hurt accuracy scores compared to many other methods.
The authors give indication that the strong performance might be due to the fact that $n$ and $m$ are tuned for the exact task at hand and not for a pruned dataset, as is done, for example, in AA.
The big downside of RA is that it ends up performing an exhaustive search over a set of options for $n$ and $m$ incurring up to $80\times$ overhead over a single training\footnote{In the original setups, the authors also used a different choice of $n$ and $m$ for the search on each task. This can be hard to do for new tasks or with less intuition for a task.}.

Fast AutoAugment (Fast AA) \cite{sungbin2019fastautoaugument} is the cheapest of the learned methods.
It is based on AA, but does not directly search for policies with strong validation performance.
Rather, it searches for augmentation policies by finding well-performing inference augmentation policies for networks trained on a split of raw, non-augmented, images.
All inference augmentations found on different splits are then joined to build a training time augmentation policy.
The intuition behind this can be summarized as follows: If a neural network trained on real data generalizes to examples augmented with some policy then this policy produces images that lie in the domain of the class, as approximated by the neural network. The augmentations therefore are class-preserving and useful.
This objective stands in contrasts to the approach followed by \advaa{}.
Fast AA tries to find augmentations that yield high accuracy when applied to validation data, while \advaa{} tries to find augmentations that yield low accuracy when applied to training data.

Finally, in an unpublished arXiv paper, Lingchen \etal \cite{lingchen2020uniformaugment} very recently suggested UniformAugment (UA), which works almost like RA.
Unlike RA, it fixes the number of augmentations to $N=2$ and drops each augmentation with a fixed probability of $0.5$.
Furthermore, the strength $m$ is sampled uniformly at random for each applied operation.

In contrast to all above methods methods, we propose \trivialaugment{} (\ta{}), an augmentation algorithm that is parameter-free like UA, but even simpler.
At the same time, \ta{} performs better than any of the comparatively cheap augmentation strategies, making it the most practical automatic augmentation method to date.

Different from all of the work discussed above, which improves final in-distribution test performance with augmentation strategies, AugMix \cite{hendrycks2020augmix} aims to improve model robustness by combining multiple augmentations in application chains, mixing their outputs, and applying a consistency loss to several augmented images. The only metric we evaluated for which AugMix was evaluated, too, is ResNet-50 performance on the ImageNet test set. Here, \ta{} outperforms AugMix.

\begin{figure}[!t]
    \centering
    \input{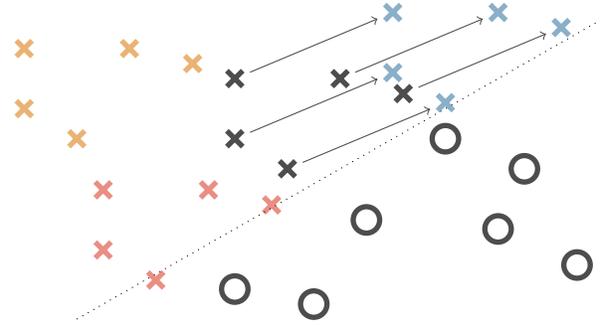}
    \caption{An exemplary visualization of a 2-D dataset with two classes, crosses and circles, separated by a decision boundary, the dotted line. The colored crosses represent deterministic augmentations of the cross class. \ta{} now uniformly samples from all crosses.}
    \label{fig:datasetvisualisation}
\end{figure}

\section{\trivialaugment{}}
\label{sec:trivialaugment}
In this section, we present the simplest augmentation algorithm we could come up with that still performs well: \emph{\trivialaugment{} (\ta{})}.
\ta{} employs the same augmentation style that previous work \cite{cubuk2020randaugment, lingchen2020uniformaugment} used:
An augmentation is defined as a function $a$ mapping an image $x$ and a discrete strength parameter $m$ to an augmented image. 
The strength parameter is not used by all augmentations, but most use it to define how strongly to distort the image. 

\ta{} works as follows.
It takes an image $x$ and a set of augmentations $\mathcal{A}$ as input. 
It then simply samples an augmentation from $\mathcal{A}$ uniformly at random and applies this augmentation to the given image $x$ with a strength $m$, sampled uniformly at random from the set of possible strengths $\{0,\dots,30\}$, and returns the augmented image.
We outline this very simple and parameter-free procedure as pseudo-code in Algorithm \ref{algo:taimp} and visualize it in Figure \ref{fig:tavisualisation}.
We emphasize that \ta{} is \emph{not} a special case of RandAugment (RA), since RA uses a \emph{fixed} optimized strength for all images while \ta{} samples this strength anew for each image.

\begin{algorithm}[]
\begin{algorithmic}[1]
\Procedure{TA}{$x$: image}

\State Sample an augmentation $a$ from $\mathcal{A}$
\State Sample a strength $m$ from $\{0,\dots,30\}$
\State Return $a(x,m)$
\EndProcedure
\end{algorithmic}
\caption{\trivialaugment{} Procedure}
\label{algo:taimp}
\end{algorithm}

While previous methods used multiple subsequent augmentations, \ta{} only applies a single augmentation to each image.
This allows viewing the distribution of the \ta{}-augmented dataset as an average of the $|\mathcal{A}|$ data distributions generated by each of the augmentations applied to the full dataset. 
In Figure \ref{fig:datasetvisualisation} we visualize this notion for deterministic augmentations without a strength parameter.
Unlike previous work, we do not generate complex distributions out of stochastic combinations of augmentation methods, but simply mean the data distributions of the augmentations applied to the given dataset.

\newcommand{\twolinecell}[2][c]{%
  \begin{tabular}[#1]{@{}c@{}}#2\end{tabular}}
  
\newcommand{\imagenetcell}[2]{\twolinecell{#1 \\ {\footnotesize (#2)}}}

\begin{table*}[!htbp]
\small
\centering
\begin{tabular}{l|cccccc|Hc}
  & Default & PBA & Fast AA & AA  &  RA & UA & \ta{} (RA) & \ta{} (Wide)\\
  \hline 
  \textbf{CIFAR-10} &&&&&&&\\
   Wide-ResNet-40-2 & 96.16 $\pm$ .08 & - & \textbf{96.4} & 96.3 & - & 96.25 & \textbf{96.62} $\pm$ .09 & 96.32 $\pm$ .05 \\ 
   Wide-ResNet-28-10 & 97.03 $\pm$ .07 & \textbf{97.4} & 97.3 & \textbf{97.4} & 97.3 & 97.33 & \textbf{97.46} $\pm$ .09 & \textbf{97.46} $\pm$ .06 \\ 
   ShakeShake-26-2x96d & 97.54 $\pm$ .07  & 98.0 & 98.0 & 98.0  & 98.0 & 98.10 & 98.05 $\pm$ .05 & \textbf{98.21} $\pm$ .06 \\ 
  PyramidNet & 97.95 $\pm$ .05 & \textbf{98.5} & \textbf{98.5} & 98.3 & \textbf{98.5} & \textbf{98.5} & 98.43 $\pm$ .04 & \textbf{98.58} $\pm$ .04 \\ 
  \hline
  \textbf{CIFAR-100} &&&&&&\\
    Wide-ResNet-40-2 & 78.42 $\pm$ .31 & - & 79.4 & 79.3 & - & 79.01 & \textbf{79.99} $\pm$ .16 & \textbf{79.86} $\pm$ .19 \\  
    Wide-ResNet-28-10 & 82.22 $\pm$ .25 & 83.3 & 82.7 & 82.9 & 83.3 & 82.82 & 83.54 $\pm$ .12 & \textbf{84.33} $\pm$ .17 \\  
    ShakeShake-26-2x96d & 83.28 $\pm$ .14 & 84.7 & 85.4 & 85.7 & - & 85.00 & 85.17 $\pm$ .23 & \textbf{86.19} $\pm$ .15\\
  \hline  
  \textbf{SVHN Core} &&&&&&\\
  Wide-ResNet-28-10 & 97.12 $\pm$ .05 & - & - & 98.0 & \textbf{98.3} & - & 98.05 $\pm$ .02 & 98.11 $\pm$ .03 \\ 
  \hline  
  \textbf{SVHN} &&&&&&\\
  Wide-ResNet-28-10 & 98.67 $\pm$ .02 & 98.9 & 98.8 & 98.9 & \textbf{99.0} & - & 98.85 $\pm$ .01 & 98.9 $\pm$ .02 \\ 
  \hline  
  \textbf{ImageNet} &&&&&&\\
  ResNet-50 & \imagenetcell{77.20 $\pm$ .32}{93.43 $\pm$ .11} & - & \imagenetcell{77.6}{93.7} & \imagenetcell{77.6}{\textbf{93.8}} & \imagenetcell{77.6}{\textbf{93.8}} & \imagenetcell{77.63}{-} &  \imagenetcell{\textbf{77.85} $\pm$ .15}{93.79 $\pm$ .07} & \imagenetcell{\textbf{78.07} $\pm$ .27}{\textbf{93.92} $\pm$ .09} \\
\end{tabular}
\caption{The average test accuracies from ten runs, besides for ImageNet, where we used five runs. The 95\% confidence interval is noted with $\pm$. The trivial \ta{} is in all benchmarks among the top-performers. The only exception is the comparison to RA\textquotesingle{s} performance on the SVHN benchmarks, but this difference was non-existent in our reimplementation in \ref{section:experiments:reproducibility}.}
\label{tab:aug_results}  
\end{table*}

\section{Experiments}
In this section, we empirically demonstrate \ta{}\textquotesingle{s} surprisingly strong performance, as well as its behaviour across many ablation settings.
In all non-ablation experiments we use either the RA augmentation space (RA), i.e., the set of augmentations and their strength parameterization from the RA paper \cite{cubuk2020randaugment}, or the wide augmentation space (Wide) for \ta{}.
We list the augmentations and their arguments for all augmentation spaces in the appendix in Table \ref{tab:searchspaces}.
We run each experiment ten times, if not stated otherwise.
In addition to the average over runs we report a confidence interval, which will contain the true mean with probability $p=95\%$, under the assumption of normal distributed accuracies.
In our code we provide a function to compute this interval. Results that lie within the confidence interval of the best performer for each task are typeset in bold font.

We evaluate our method on five different datasets.
i) CIFAR-10 and CIFAR-100 \cite{krizhevsky2009learningcifar} are standard datasets for image classification and each contain 50K training images. We trained Wide-ResNets \cite{wrn} as well as a ShakeShake model \cite{gastaldi2017shakeshake}. We follow previous work \cite{cubuk2019autoaugment,cubuk2020randaugment} with our setup. 
ii) SVHN \cite{svhn} consists of images of house numbers.
It comes with a core set of 73K training images, but offers an additional 531K simpler images as extension of the dataset. We perform experiments with and without the additional images on a Wide-ResNet-28-10.
iii) Finally, we perform experiments on ImageNet, a very large image classification corpus with 1000 classes and over 1.2 million images. This experiment is particularly interesting, since it was shown previously that there are augmentations, such as cutout, that do not generalize well to ImageNet. We train a ResNet-50 \cite{he2016deep} following the setup of \cite{cubuk2019autoaugment}. We use warmup and 32 workers due to cluster limitations, which is less than \cite{cubuk2019autoaugment}. We scale the learning rate appropriately.
See Appendix \ref{section:trainingsettings} for more details.

\subsection{Comparison to State-of-the-Art}
\label{section:comparison}
It is non-trivial to compare automatic augmentation methods fairly.
We therefore compare our method with the previous state-of-the-art in three different setups.

In Section \ref{section:comparison:fixedtrainingsetup}, we follow the majority of previous work \cite{cubuk2019autoaugment,cubuk2020randaugment,ho2019populationbasedaugmentation,sungbin2019fastautoaugument,lingchen2020uniformaugment} and perform a comparison with other methods that use the same model and training pipeline.
This setup allows for different search costs of different methods and compares methods with the same inference and training costs.

In Section \ref{section:experiments:reproducibility}, we compare in a similar way as above, but against reproductions of other methods in our codebase.
This avoids confounding factors, making sure that the methods, and not setup details, explain the differences between results.
We reproduced a total of four other methods in our codebase, including the cheapest three previous methods. 

In Section \ref{section:comparison:totalcompute}, we compare the total cost of each method, both search and model training, with the final accuracy. This comparison has the upside that it can consider work with different pipelines and models more fairly.

\subsubsection{Comparison to Published Results}
\label{section:comparison:fixedtrainingsetup}
In Table \ref{tab:aug_results}, we compare \ta{} to all methods that used the setup of AutoAugment \cite{cubuk2019autoaugment} or a very similar setup in terms of hyper-parameters, number of epochs and models.

\ta{} performs as well or better than previous methods in almost all tasks. 
The SVHN datasets are the only exception, with RA performing somewhat better.
This might, however, be due to our training pipeline, since, as we show in Section \ref{section:experiments:reproducibility}, we were not able to reproduce RA\textquotesingle{s} performance for SVHN Core with our pipeline and the original training pipeline is not available.

For ImageNet, \ta{} outperformed all other methods in terms of both top-1 accuracy and top-5 accuracy.
We used an image width of 244 like RA \cite{cubuk2020randaugment}, but even with a lower width of 224 (as was used for AA \cite{cubuk2019autoaugment}), \ta{} outperformed the previously best methods (with a 77.97 $\pm$ .21 top-1 accuracy and 93.98 $\pm$ .07 top-5 accuracy; not listed in the table).

In this comparison, we cannot compare to all previous methods, since some use different setups.
The best-known setup we had to leave out is \advaa{}. Therefore, we perform an extra set of experiments following its setup closely.

\begin{table}[]
\small
    \centering
\begin{tabular}{l|c|c}
  & Adv. AA & \ta{} (Wide)\\
  \hline 
  \textbf{CIFAR-10} &\\
   Wide-ResNet-28-10 &\textbf{98.10} $\pm$ .15 & \textbf{98.04} $\pm$ .06  \\ 
   ShakeShake-26-2x96d  & \textbf{98.15} $\pm$ .12 & \textbf{98.12} $\pm$ .12 \\ 
  \hline
  \textbf{CIFAR-100} &&\\
    Wide-ResNet-28-10 & \textbf{84.51} $\pm$ .18 & \textbf{84.62} $\pm$ .14\\  
    ShakeShake-26-2x96d &  \textbf{85.90} $\pm$ .15 & \textbf{86.02} $\pm$ .13\\
\end{tabular}
    \caption{A comparison of \ta{} with \advaa{} in the augmented batch setting on a Wide-ResNet-28-10. We report the average over five runs.}
    \label{tab:adv_aa_comparison}
\end{table}

\advaa{} uses eight times the compute for its final training compared to other methods and therefore has a significant advantage compared to other methods.
\advaa{} is based on batch augmentation \cite{Hoffer20_AugmentYourBatch}, where a set of workers in a data parallel setting each compute gradients with respect to the same batch of examples, but apply different augmentations to the images in it.
We re-created this setup, including all hyper-parameters and batch augmentation, for \ta{}.
In Table \ref{tab:adv_aa_comparison}, we compare \ta{} with \advaa{} with a Wide-ResNet-28-10 and a ShakeShake-26-2x96d for both CIFAR-10 and CIFAR-100.
We show that TA\textquotesingle{s} trivial uniform sampling of a single augmentation achieves the same performance as their complex (and unavailable) reinforcement learning pipeline.

We conclude from this section that, for almost all considered benchmarks across datasets, models and even the way augmentations are applied, \ta{} is among the top-performing methods.

\subsubsection{Comparison of Reproduced Results in a Fixed Training Setup}
\label{section:experiments:reproducibility}
While in the previous section, we tried to mitigate confounding factors by comparing results obtained with very similar setups with each other, in this section, we go one step further. We reproduce the results of four methods and compare our baseline method with these reproductions in order to yield a true apples-to-apples comparison.

As we present a very cheap and simple augmentation method we picked RA, Fast AA and UA as other cheap and simple augmentation methods for our comparison.
Additionally, we compare to AA, as an important, common baseline.
Moreover, for all of these methods relevant information for reproduction was published\footnote{See Table \ref{tab:reproducibilityofpapers} in the appendix for an overview of the published materials of different methods}.

For RA, AA and Fast AA we used the published policies and did not search for an augmentation policy from scratch.
We based both our RA and AA implementations on a public codebase\footnote{\url{https://github.com/tensorflow/models/tree/fd34f711f319d8c6fe85110d9df6e1784cc5a6ca/research/autoaugment}} by the authors of both RA and AA that implements AA for the CIFAR datasets.
Likewise, for Fast AA we based our implementation on a public codebase.
No code is published for UA, and there are multiple hyper-parameters missing in the paper; in these cases, we used the hyper-parameters from RA.
For our reproduction of UA, we also adopted the same discretization of the augmentation strengths into 31 values used by the other methods.
In addition to the original augmentation space of UA we also perform experiments with the RA augmentation space.

We reran experiments for CIFAR-10, CIFAR-100 and SVHN Core, and present the results in Table \ref{tab:cifarsvhnreimplementation}.
For each method we ran the benchmarks included in the original work.
Generally, we could reproduce most results or even improve upon published results.
The only severe exception is RA for which we tried multiple changes to the setup, but were not able to reach their published scores -- neither for CIFAR nor for SVHN Core.

In this evaluation, \ta{} (Wide) performed best across all methods for each benchmark with a Wide-Resnet-28-10, and \ta{} (RA) performed best for both Wide-Resnet-40-2 benchmarks.

In addition to the reproductions of published policies, we applied RandAugment to the Wide-ResNet-40-2 on CIFAR-10, which was originally not considered in the RA paper.
We therefore had to search for a policy first.
Depending on the task, Cubuk \etal \cite{cubuk2020randaugment} considered different subsets of the full range of the augmentation strengths $M \subset \{1,\dots,30\}$ and the number of consecutive augmentations $N \subset \{1,\dots,3\}$.
In order to avoid missing the best candidates and to not require human intuition we searched on all 90 resulting combinations of RA\textquotesingle{s} parameters.
We split up a validation set of 10000 examples like in the original RandAugment method to evaluate the settings. We then picked the best setting and compared it to \ta{}.
Table \ref{tab:bruteforcerandaugment} clearly shows that \ta{} performs better than the costly RA setup, even though the RA setup in total required 91 full trainings, compared to a single training for \ta{}.

Finally, we consider three more evaluations in the appendix: (i) We show that TA performs comparably or better on the same augmentation space with other automatic augmentation methods (see Appendix \ref{sec:comwsameaugs}), (ii) we show that TA generalizes to more peculiar datasets (see Appendix \ref{sec:specialdatasets}) and (iii) we show TA\textquotesingle{s} effectiveness with the EfficientNet Architecture \cite{tan2019efficientnet} (see Appendix \ref{sec:efficientneteval}).

\begin{table}[h]
\small
\centering
    \begin{subtable}{.5\textwidth}
    \centering
     \begin{tabular}{c|cc}
            WRN-28-10 & CIFAR-10 & CIFAR-100 \\
            \hline
         AA & 97.31 $\pm$ .22 \textcolor{gray}{(-.09)} & 82.91 $\pm$ .41  \textcolor{gray}{(+.01)}  \\ 
         FAA & \textbf{97.43} $\pm$ .09 \textcolor{gray}{(+.13)} & 83.27 $\pm$ .13 \textcolor{gray}{(+.57)}\\ 
         RA & 97.12 $\pm$ .14 \textcolor{gray}{(-.18)} & 83.1 $\pm$ .32  \textcolor{gray}{(-.20)}\\ 
         UA (UA) & \textbf{97.46} $\pm$ .14 \textcolor{gray}{(+.13)} & 83.08 $\pm$ .27 \textcolor{gray}{(+.26)} \\ 
         UA (RA) & \textbf{97.44} $\pm$ .09 & 83.36 $\pm$ .18\\ 
         TA (RA) & \textbf{97.46} $\pm$ .09 & 83.54 $\pm$ .12\\
         TA (Wide) & \textbf{97.46} $\pm$ .06 & \textbf{84.33} $\pm$ .17
    \end{tabular}
    \subcaption{}
    \label{tab:cifar10reimplementation}
    \end{subtable}
    
    \begin{subtable}{.5\textwidth}
    \centering
     \begin{tabular}{c|cc}
            WRN-40-2 & CIFAR-10 & CIFAR-100 \\
            \hline
         AA & 96.38 $\pm$ .10 \textcolor{gray}{(+.08)} & 79.66 $\pm$ .17  \textcolor{gray}{(+.36)}  \\ 
         FAA & 96.39 $\pm$ .06 \textcolor{gray}{(-.01)} & 79.79 $\pm$ .21 \textcolor{gray}{(+.39)}\\ 
         UA (UA) & 96.42 $\pm$ .04 \textcolor{gray}{(+.17)} & 79.74 $\pm$ .15 \textcolor{gray}{(+.73)} \\ 
         UA (RA) & 96.45 $\pm$ .06 & \textbf{79.95} $\pm$ .20\\ 
         TA (RA) & \textbf{96.62} $\pm$ .09 & \textbf{79.99} $\pm$ .16\\ 
         TA (Wide) & 96.32 $\pm$ .05 & \textbf{79.86} $\pm$ .19\\ 
    \end{tabular}
    \subcaption{}
    \label{tab:cifar100reimplementation}
    \end{subtable}
    
    \begin{subtable}{.5\textwidth}
    \centering
    \begin{tabular}{c|c}
            WRN-28-10 & SVHN Core \\
            \hline
         AA & 97.99 $\pm$ .06 \textcolor{gray}{(-.01)} \\ 
         RA & 98.06 $\pm$ .04 \textcolor{gray}{(-.24)} \\ 
         \ta{} (RA) & 98.05 $\pm$ .02 \\
         \ta{} (Wide) & \textbf{98.11} $\pm$ .03 \\
    \end{tabular}
    \subcaption{}
    \label{tab:svhnreimplementation}
     
    \end{subtable}
    \caption{A reproduction of the results of previous work with a Wide-ResNet-28-10 on CIFAR (a) and SVHN Core (c), and with a Wide-ResNet-40-2 on CIFAR (b). We report the relative performance difference to the published results in parentheses.} 
    \label{tab:cifarsvhnreimplementation}
\end{table}

\begin{table}[]
\small
    \centering
    \begin{tabular}{l|c}
         Method & Acc. \\
         \hline
         Brute-Force RA & 96.42 $\pm$ .09 \\
         \ta{} (RA) & \textbf{96.62 $\pm$ .09}\\
    \end{tabular}
    \caption{Average over ten runs on CIFAR-10 with a Wide-ResNet-40-2. \ta{} performs better than the over 80-times more expensive exhaustive search over RA\textquotesingle{s} parameters.
}
    \label{tab:bruteforcerandaugment}
\end{table}

\begin{figure}[b]
    \centering
    \definecolor{color0}{rgb}{0.587254901960784,0.793137254901961,0.757843137254902}
\definecolor{color1}{rgb}{0.962745098039216,0.962745098039216,0.73921568627451}
\definecolor{color2}{rgb}{0.756862745098039,0.745098039215686,0.83921568627451}
\definecolor{color3}{rgb}{0.917156862745098,0.555392156862745,0.51421568627451}
\definecolor{color4}{rgb}{0.542647058823529,0.686764705882353,0.786764705882353}
\definecolor{color5}{rgb}{0.916176470588235,0.701470588235294,0.460294117647059}
\definecolor{color6}{rgb}{0.686764705882353,0.813235294117647,0.469117647058823}
\definecolor{color7}{rgb}{0.965196078431373,0.826960784313726,0.897549019607843}
\definecolor{color8}{rgb}{0.708333333333333,0.531862745098039,0.711274509803922}
\definecolor{color9}{rgb}{0.811764705882353,0.902941176470588,0.791176470588235}
\definecolor{color10}{rgb}{0.929411764705882,0.876470588235294,0.505882352941176}

\pgfplotscreateplotcyclelist{waldorfcolors}{%
color0,color1,color2,color3,color4,color5,color6,color7,color8,color9,color10}

\begin{tikzpicture}
\pgfplotsset{
        cycle list/Dark2,
        cycle multiindex* list={
            [3 of]mark list\nextlist
            Dark2\nextlist
        },
    }
\small
\begin{axis}[
    xmode=log,
    log ticks with fixed point,
    xlabel=GPU hours,
    ylabel=Accuracy (\%),
    width=.5\textwidth,height=.4\textwidth,
    legend columns=2,
    legend style={at={(0.324,0.045)},anchor=south west},
]

\addplot+[only marks] table{
1602 79.3
1606 82.9
1641 85.7
};
\addlegendentry{AA};

\addplot+ table {
4.4 79.4
11.2 82.7
83 85.4
};
\addlegendentry{Fast AA};

\addplot+ table {
4 78.3
25.25 83.3
};
\addlegendentry{RA};

\addplot+[only marks] table {
125 84.7
158 85.93
};
\addlegendentry{AWS}

\addplot+[only marks] table {
2.2 79.01
5.6 82.82
41.5 85.00
};
\addlegendentry{UA}

\addplot+[only marks] table {
161 85.84
338 86.04
};
\addlegendentry{AWS x8}

\addplot+[mark=*,color=blue!80] table {
2.2 79.99
5.6 83.54
41.5 85.15
};
\addlegendentry{TA (RA)}

\addplot+[only marks] table {
44.8 84.51
222 85.9
};
\addlegendentry{\advaa{} x8}
\addplot+[color=blue!80] table {
2.2 79.86
5.6 84.33
41.5 86.2
};
\addlegendentry{TA (Wide)}

\addplot+[color=blue!80] table {
44.8 84.62
222 86.02
};
\addlegendentry{TA (Wide) x8}

\end{axis}
\end{tikzpicture}
    \caption{Comparison of the final test accuracy on CIFAR-100 in comparison to RTX2080ti GPU-hours compute invested for \textit{augmentation search and final model training} across a set of models. Methods marked with \emph{x8} use batch augmentations\cite{Hoffer20_AugmentYourBatch}.}
    \label{fig:performancepercompute}
\end{figure}
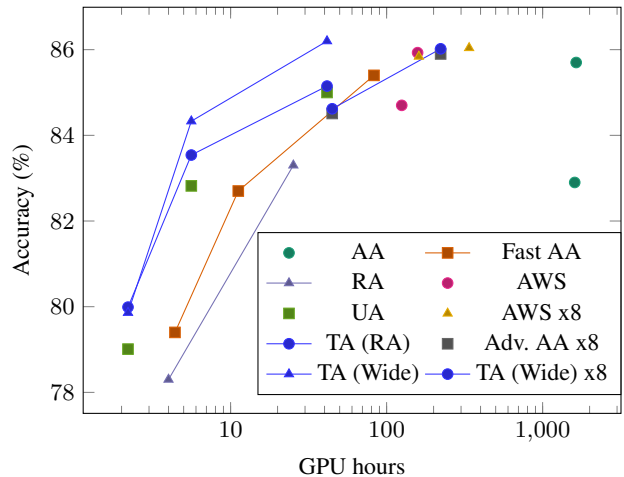

\subsubsection{Comparison by Total Compute Costs}
\label{section:comparison:totalcompute}
In the previous sections, we compared different augmentation methods for a fixed training setup.
We now consider the other extreme, comparing all methods across models and setups by their compute requirements.

In Figure \ref{fig:performancepercompute}, we plot this comparison for many CIFAR-100 setups across the literature.
The question this plot answers is: given some compute budget, what method should we choose for the best final accuracy?
For this plot, we used the accuracy numbers published in the literature and estimated the compute costs in RTX 2080 Ti GPU-hours.
See Appendix \ref{sec:computecostapproximation} for a detailed account of the information used to calculate the compute cost approximations for all setups.
We had to restrict the set of models we considered to the set of models for which we know from our experiments how expensive they are to run, namely all Wide-ResNet setups and the ShakeShake-26-2x96d.
We tried to be as conservative as possible regarding the compute requirements of other methods, to not give \ta{} an unfair advantage.

In the figure, for all considered budgets, \ta{} and its variant with augmented batch (\ta{} x8) perform among the best methods.
\ta{} also has a clear benefit compared to the popular cheap methods Fast AA and RA for all compute budgets; finally, it is dramatically cheaper than AA.

\begin{table}[]
\small
    \centering
    \begin{tabular}{c|cc}
    Augmentation space & SVHN Core & CIFAR-10\\
    \hline
    Full & 97.63 $\pm$ .06 & 97.24 $\pm$ .03 \\ 
    AA & 98.04 $\pm$ .02 & 97.47 $\pm$ .11 \\ 
    AA - $\{$Invert$\}$ & 97.97 $\pm$ .08 & \textbf{97.55} $\pm$ .06 \\ 
    RA &  98.05 $\pm$ .02 & 97.46 $\pm$ .09 \\ 
    Wide &  \textbf{98.11} $\pm$ .03 & 97.46 $\pm$ .06 \\ 
    UA & 98.06 $\pm$ .04 & 97.42 $\pm$ .07 \\ 
    OHL & \textbf{98.10} $\pm$ .02 & 97.45 $\pm$ .05 \\ 
    \end{tabular}
    \caption{Evaluation of \ta{} on SVHN Core and CIFAR-10 with a set of 7 different augmentation spaces. Note that $\text{RA} = \text{AA} - \{\text{SamplePairing}, \text{Invert}, \text{Cutout}\}$ and $\text{UA} = \text{AA} - \{\text{SamplePairing}\}$.}
    \label{tab:impact_of_search_space}
\end{table}

\subsection{Understanding the Minimal Requirements of \ta{}}
While so far, we have demonstrated that in many circumstances TA\textquotesingle{s} approach of only using a single augmentation per image is enough or yields even better performance than more complicated methods,
in this section we will dissect other properties of \ta{}.

We first analyse how \ta{} behaves across augmentation spaces from the literature.
We then look at its performance after we apply random changes to its augmentation space.
Finally, we consider sets of different augmentation strengths from which \ta{} samples.

\subsubsection{\ta{} with Different Hand-Picked Augmentation Spaces}
For this evaluation, we carefully reimplemented the augmentation spaces of AA, UA and OHL, besides the one of RA.
Additionally, we consider a larger augmentation space (Full), which is a super set of AA, and additionally contains a blur, a smooth, a horizontal and a vertical flip.
Especially the vertical flip is likely not useful for very many classification tasks.
See Table \ref{tab:searchspaces} in the appendix for an overview of the augmentation spaces.

Table \ref{tab:impact_of_search_space} indeed shows that \ta{} performs worse on the full augmentation space than on all other augmentation spaces for a Wide-ResNet-28-10 on both SVHN Core and CIFAR-10.

We also included another augmentation space not considered in the previous literature: a variant of the AA augmentation space, where we removed the extreme invert operation, which maps each pixel $x$ to $255-x$.
We can see that this augmentation space performs very well for CIFAR-10, but not great for SVHN Core.
This aligns well with observations made by earlier work, indicating that the invert augmentation fosters generalization on SVHN, but not on the other datasets\cite{cubuk2019autoaugment}.
A peculiarity of the OHL augmentation space is that it only uses three strengths, unlike all other methods which consider 31 strengths. 
Interestingly, this is not harmful and OHL yields the best score for SVHN Core.

We can see that the performance of \ta{} is rather stable between augmentation spaces, but still there seems to be room for improvement by a more sophisticated method to choose the augmentation space for \ta{} depending on the task.

\begin{figure}[]
\small
    \centering
    \input{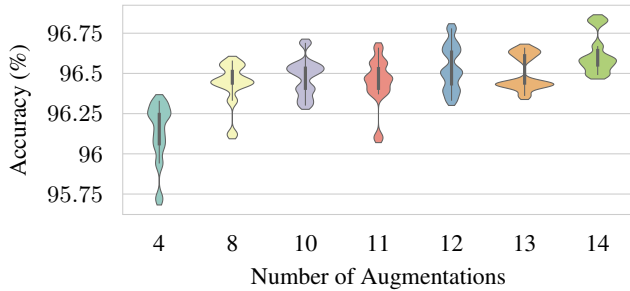}
    \caption{The performance of WRN-40-2 models depending on the size of sampled subsets of the RA augmentations on CIFAR-10. We performed 10 evaluations per subset size.}
    \label{fig:subsampledta}
\end{figure}

\subsubsection{\ta{}\textquotesingle{s} Behavior With Randomly Pruned Augmentation Spaces}
While we assessed performance with different hand-crafted augmentation spaces above, now we want to analyze how performance is impacted if we only use random subsets of the 14 augmentations in the RA augmentation space (which we used in the other experiments unless otherwise stated). 

In Figure \ref{fig:subsampledta}, we analyze the performance and its variance for multiple augmentation subset sizes for a Wide-ResNet-40-2 on CIFAR-10.
We performed 10 evaluations per sample size, where in each evaluation we picked a random sample of augmentations.
While performance decreases as fewer and fewer augmentations are considered, we can see that it drops very slowly.
We can throw away 4 of 14 augmentations and still obtain performance close to the original performance.
Another trend is that with fewer augmentations the variance increases.
This is likely due to the randomness of the subset choice per run, which increases for smaller subsets.

\subsubsection{The Impact of the Set of Strengths on \ta{}\textquotesingle{s} Performance}
Before, we mostly considered the impact of different sets of augmentations; now we consider the other component of the augmentation space: the set of strengths.

In Table \ref{tab:differentstrengthsets}, we analyze the performance of \ta{} with a Wide-ResNet-28-10 and different subsets of the original set of possible strengths $\{0,\dots,30\}$ on the RA augmentation space.
We can see that the CIFAR-10 setup seems to be relatively agnostic to the set of strengths. 
Performance on CIFAR-100, on the other hand, is very negatively impacted by choosing the subset $\{30\}$.
In general, performance improves on CIFAR-100 with larger sets.
For SVHN Core, the opposite is the case: performance improves when only considering $\{30\}$.
A reason for this could be that the majority of the augmentations are color based and changing the colors of a single-color background and a single-color number drastically, still in most cases yields valid house numbers.

Another observation we made is that it does not matter so much for any setup whether we reduce to three or just two augmentation strengths, compared to all 31.
This seems to point towards the importance of a mixture of strong and weak augmentations. At the same time three different strengths, compared to 31, seem to be enough for these settings.

\begin{table}[]
\small
    \centering
    \begin{tabular}{l|ccc}
        Strengths& CIFAR-10 & CIFAR-100 & SVHN Core\\
        \hline
         $\{30\}$ & \textbf{97.45} $\pm$ .05 & 82.98 $\pm$ .22 & \textbf{98.16} $\pm$ .03 \\
         $\{0,30\}$ & \textbf{97.51} $\pm$ .08 & \textbf{83.46} $\pm$ .10 & 98.02 $\pm$ .02 \\ 
         $\{0,15,30\}$ & \textbf{97.46} $\pm$ .06 & \textbf{83.43} $\pm$ .24 & 98.04 $\pm$ .03 \\ 
         $\{0,\dots,30\}$ & \textbf{97.46} $\pm$ .09 & \textbf{83.54} $\pm$ .12 & 98.05 $\pm$ .02 \\
    \end{tabular}
    \caption{A comparison of the performance of \ta{} (RA) on different datasets with a Wide-ResNet-28-10 using different subsets of strengths.}
    \label{tab:differentstrengthsets}
\end{table}

\section{Automatic Augmentation Methods in \\Practice}
While there are many expensive or hard to reproduce automatic augmentation methods, it is important that augmentation methods are practical: the impact of automatic augmentation methods unfolds in the application to new setups and problems.
We evaluated many different settings and augmentation methods and we would like to pass on the gained knowledge.

First, we have compiled a short summary of learnings for the application of augmentation methods in Appendix \ref{sec:applicationrecommendation}.

\definecolor{codegreen}{rgb}{0.2,0.6,0.2}
\definecolor{codegray}{rgb}{0.5,0.5,0.5}
\definecolor{codepurple}{rgb}{0.58,0,0.82}
\definecolor{backcolour}{rgb}{0.95,0.95,0.92}

\lstdefinestyle{mystyle}{
    backgroundcolor=\color{white},   
    commentstyle=\color{codegreen},
    keywordstyle=\color{magenta},
    numberstyle=\tiny\color{codegray},
    stringstyle=\color{codepurple},
    emph={RandAugment,TrivialAugment,UniformAugment},
    emphstyle={\color{codegreen}},
    basicstyle=\ttfamily,
    breakatwhitespace=false,         
    breaklines=true,                 
    captionpos=b,                    
    keepspaces=true,                 
    numbers=none,                    
    numbersep=5pt,                  
    showspaces=false,                
    showstringspaces=false,
    showtabs=false,                  
    tabsize=2
}
\lstset{style=mystyle}

Second, in addition to our full codebase, we provide a simple one-file python library that implements the more practical augmentation methods: RA, UA and \ta{}.
It even allows choosing from all augmentation spaces considered in this work.
For example, to get an image augmenter for TA and transform a PIL image \lstinline{img}, one can call  
\begin{lstlisting}[language=Python, frame=single, numbers=left]
aug = TrivialAugment(n,m)
augmented_img = aug(img)
\end{lstlisting}

\section{Best Practices Proposal for Research}
We found that it is difficult to reimplement many of the published methods, see Table \ref{tab:reproducibilityofpapers} in the appendix.
We also found that many methods performed similarly to the simple \ta{} baseline, when we follow their setup.
Here, we compile a short bullet point list of best practices we believe are important for sustainable research in this field.

\begin{itemize}
\item Share code as much as possible for easy entry of beginners and to make sure that setups are similar across papers. Otherwise, differences between the actual implementation and its description in the paper can impair reproducibility.
\item Compare fairly to other methods and baselines with the same setup, train budget and augmentation space, or reproduce results of previous methods in your setup and mention differences.
\item Report confidence intervals to discern ``outperforming" from ``performing comparably".
\end{itemize}

\section{Limitations}
While we could not find settings where TA failed for image classification, we found that TA does not work out-of-the-box for object detection setups and also needs tuning to work for this task. So far, we can only wholeheartedly recommend the use of TA for image classification; its application to other computer vision tasks requires further study.

\section{Conclusion}
Most of the approaches considered as automatic augmentation methods are complicated.
In this work, we presented \ta{}, a very simple augmentation algorithm from which we can learn three main things.

First, \ta{} teaches us about a crucial baseline missing for automatic augmentation methods.

Second, \ta{} teaches us to never overlook the simplest solutions.
There are a lot of complicated methods to automatically find augmentation policies, but the simplest method was so-far overlooked, even though it performs comparably or better.

Third, randomness in the chosen strengths appears to be very important for good performance.

\section*{Acknowledgements}
We want to thank Ildoo Kim for his open-source codebase which ours forks from, and the reviewers for their insightful comments. We acknowledge funding by the Robert Bosch GmbH and the European Research Council (ERC) under the European Union Horizon 2020 research and innovation programme through grant no. 716721. 

{\small
\bibliographystyle{ieee_fullname}
\bibliography{lib,strings,proc,main}

\begin{thebibliography}{10}\itemsep=-1pt

\bibitem{cubuk2019autoaugment}
Ekin~D Cubuk, Barret Zoph, Dandelion Mane, Vijay Vasudevan, and Quoc~V Le.
\newblock Autoaugment: Learning augmentation strategies from data.
\newblock In {\em Proceedings of the IEEE/CVF Conference on Computer Vision and
  Pattern Recognition}, pages 113--123, 2019.

\bibitem{cubuk2020randaugment}
Ekin~D Cubuk, Barret Zoph, Jonathon Shlens, and Quoc~V Le.
\newblock Randaugment: Practical automated data augmentation with a reduced
  search space.
\newblock In {\em Proceedings of the IEEE/CVF Conference on Computer Vision and
  Pattern Recognition Workshops}, pages 702--703, 2020.

\bibitem{devries2017improved}
Terrance DeVries and Graham~W. Taylor.
\newblock Improved regularization of convolutional neural networks with cutout,
  2017.

\bibitem{fadaee2017dataaugformt}
Marzieh Fadaee, Arianna Bisazza, and Christof Monz.
\newblock Data augmentation for low-resource neural machine translation.
\newblock In {\em Proceedings of the 55th Annual Meeting of the Association for
  Computational Linguistics (Volume 2: Short Papers)}, pages 567--573, 2017.

\bibitem{gastaldi2017shakeshake}
Xavier Gastaldi.
\newblock Shake-shake regularization, 2017.

\bibitem{Detectron2018}
Ross Girshick, Ilija Radosavovic, Georgia Gkioxari, Piotr Doll\'{a}r, and
  Kaiming He.
\newblock Detectron.
\newblock \url{https://github.com/facebookresearch/detectron}, 2018.

\bibitem{he2016deep}
Kaiming He, Xiangyu Zhang, Shaoqing Ren, and Jian Sun.
\newblock Deep residual learning for image recognition.
\newblock In {\em Proceedings of the IEEE conference on computer vision and
  pattern recognition}, pages 770--778, 2016.

\bibitem{hendrycks2020augmix}
Dan Hendrycks, Norman Mu, Ekin~D. Cubuk, Barret Zoph, Justin Gilmer, and Balaji
  Lakshminarayanan.
\newblock {AugMix}: A simple data processing method to improve robustness and
  uncertainty.
\newblock {\em Proceedings of the International Conference on Learning
  Representations (ICLR)}, 2020.

\bibitem{ho2019populationbasedaugmentation}
Daniel Ho, Eric Liang, Xi Chen, Ion Stoica, and Pieter Abbeel.
\newblock Population based augmentation: Efficient learning of augmentation
  policy schedules.
\newblock In {\em International Conference on Machine Learning}, pages
  2731--2741. PMLR, 2019.

\bibitem{Hoffer20_AugmentYourBatch}
Elad Hoffer, Tal Ben-Nun, Itay Hubara, Niv Giladi, Torsten Hoefler, and Daniel
  Soudry.
\newblock Augment your batch: Improving generalization through instance
  repetition.
\newblock In {\em Proceedings of the IEEE/CVF Conference on Computer Vision and
  Pattern Recognition (CVPR)}, June 2020.

\bibitem{krizhevsky2009learningcifar}
Alex Krizhevsky et~al.
\newblock Learning multiple layers of features from tiny images.
\newblock 2009.

\bibitem{krizhevsky-nips12}
A. Krizhevsky, I. Sutskever, and G. Hinton.
\newblock {ImageNet} classification with deep convolutional neural networks.
\newblock In P. Bartlett, F. Pereira, C. Burges, L. Bottou, and K. Weinberger,
  editors, {\em Proceedings of the 26th International Conference on Advances in
  Neural Information Processing Systems ({N}eur{IPS}'12)}, pages 1097--1105,
  2012.

\bibitem{sungbin2019fastautoaugument}
Sungbin Lim, Ildoo Kim, Taesup Kim, Chiheon Kim, and Sungwoong Kim.
\newblock Fast autoaugment.
\newblock In H. Wallach, H. Larochelle, A. Beygelzimer, F. d\textquotesingle
  Alch\'{e}-Buc, E. Fox, and R. Garnett, editors, {\em Advances in Neural
  Information Processing Systems}, volume~32, pages 6665--6675. Curran
  Associates, Inc., 2019.

\bibitem{Lin2019ohlautoaug}
Chen Lin, Minghao Guo, Chuming Li, Xin Yuan, Wei Wu, Junjie Yan, Dahua Lin, and
  Wanli Ouyang.
\newblock Online hyper-parameter learning for auto-augmentation strategy.
\newblock In {\em Proceedings of the IEEE/CVF International Conference on
  Computer Vision (ICCV)}, October 2019.

\bibitem{lingchen2020uniformaugment}
Tom~Ching LingChen, Ava Khonsari, Amirreza Lashkari, Mina~Rafi Nazari,
  Jaspreet~Singh Sambee, and Mario~A. Nascimento.
\newblock Uniformaugment: A search-free probabilistic data augmentation
  approach, 2020.

\bibitem{loshchilov-iclr17aa}
I. Loshchilov and F. Hutter.
\newblock Sgdr: Stochastic gradient descent with warm restarts.
\newblock In {\em Proceedings of the International Conference on Learning
  Representations (ICLR'17)}, 2017.

\bibitem{svhn}
Yuval Netzer, Tao Wang, Adam Coates, Alessandro Bissacco, Bo Wu, and Andrew~Y.
  Ng.
\newblock Reading digits in natural images with unsupervised feature learning.
\newblock In {\em NIPS Workshop on Deep Learning and Unsupervised Feature
  Learning 2011}, 2011.

\bibitem{tan2019efficientnet}
Mingxing Tan and Quoc Le.
\newblock Efficientnet: Rethinking model scaling for convolutional neural
  networks.
\newblock In {\em International Conference on Machine Learning}, pages
  6105--6114. PMLR, 2019.

\bibitem{tian2020awsimprovingautoaug}
Keyu Tian, Chen Lin, Ming Sun, Luping Zhou, Junjie Yan, and Wanli Ouyang.
\newblock Improving auto-augment via augmentation-wise weight sharing.
\newblock {\em Advances in Neural Information Processing Systems}, 33, 2020.

\bibitem{UDAdataaugmentationsemisupervised}
Qizhe Xie, Zihang Dai, Eduard Hovy, Thang Luong, and Quoc Le.
\newblock Unsupervised data augmentation for consistency training.
\newblock In H. Larochelle, M. Ranzato, R. Hadsell, M.~F. Balcan, and H. Lin,
  editors, {\em Advances in Neural Information Processing Systems}, volume~33,
  pages 6256--6268. Curran Associates, Inc., 2020.

\bibitem{wrn}
Sergey Zagoruyko and Nikos Komodakis.
\newblock Wide residual networks.
\newblock In {\em In Proceedings of {BMCV}'16}, 2016.

\bibitem{zhang2020adversarialaa}
Xinyu Zhang, Qiang Wang, Jian Zhang, and Zhao Zhong.
\newblock Adversarial autoaugment.
\newblock In {\em International Conference on Learning Representations}, 2020.

\bibitem{zoph2019objectdetectionaugment}
Barret Zoph, Ekin~D Cubuk, Golnaz Ghiasi, Tsung-Yi Lin, Jonathon Shlens, and
  Quoc~V Le.
\newblock Learning data augmentation strategies for object detection.
\newblock In {\em European Conference on Computer Vision}, pages 566--583.
  Springer, 2020.

\end{thebibliography}
}

\newpage
\appendix

\section{Training Settings}
\label{section:trainingsettings}
For all setups we normalize the images by training set mean and standard deviation after the application of all augmentations, besides a final cutout, if applicable.
\subsection{CIFAR}
Following previous work we apply the vertical flip and the pad-and-crop augmentations and finally a 16 pixel cutout \cite{devries2017improved} after \ta{} or generally any augmentation method.
We trained Wide-ResNet models \cite{wrn} in the Wide-ResNet-40-2 and the larger Wide-ResNet-28-10 settings.
We trained these models for 200 epochs using SGD with Nesterov Momentum and a learning rate of 0.1, a batch size of 128, a 5e-4 weight decay, cosine learning rate decay \cite{loshchilov-iclr17aa}.

We trained ShakeShake-26-2x96d for 1600 epochs using SGD with Nesterov Momentum, a learning rate of 0.01, a batch size of 128, 1e-3 weight decay and a cosine learning rate decay.

For the augmented batch setups we followed Zhang \etal \cite{zhang2020adversarialaa}. We used the settings above for the Wide-ResNet-28-10 evaluations. And like Zhang slightly different settings for ShakeShake. We use 600 epochs, with a 0.2 learning rate and a 1e-4 weight decay.

\subsection{SVHN}
Unlike for CIFAR we do not apply extra augmentations for SVHN, besides a final 16 pixel cutout \cite{devries2017improved}.
For the full dataset we trained for 160 epochs using SGD with Nesterov Momentum of 0.9, a learning rate of 0.005, a batch size of 128, a 1e-3 weight decay and cosine learning rate decay.
For SVHN Core we train with the same settings, except that we trained for 200 epochs and used a larger weight decay of 5e-3.

\subsection{ImageNet}
Like for the other datasets we performed the standard augmentations of the dataset after the learned augmentations. That is we performed a randomly resized crop and scales between 0.08 and 1.0 using bicubic interpolation. We augmented with horizontal flips, applied a color jitter with brightness, contrast and saturation strength set to 0.4 and we applied lighting noise with an alpha of 0.1.

We trained on Imagenet with a ResNet-50 \cite{he2016deep} and followed the setup of AA \cite{cubuk2019autoaugment}. We train for 270 epochs with a batch size of 2048 distributed among 32 workers.
We use image crops of height 224 considered both a 244 width of the images, like RA, and a 224 width, like AA. 
The initial learning rate of 0.1 is scaled proportional to the batch size divided by 256. As learning rate schedule we apply a step-wise 10-fold reduction after 90, 180 and 240 epochs with a linear warmup of factor 4 over the first 3 epochs.
We use Nesterov Momenutm with a momentum parameter of 0.9 and a weight decay of 1e-4.

Unlike \cite{cubuk2019autoaugment} we only use 32 instead of 64 workers out of cluster limitations and scale the learning rate accordingly.

\definecolor{Added}{rgb}{1.0, 1.0, 0.13}
\begin{table}[h]
\centering
\scalebox{.85}{
    \begin{tabular}{c|c  c|c}
            PIL operation & range &  PIL operation & range \\
            {{identity}}& - & {{auto\_contrast}}& - \\
            {{equalize}}& - & rotate & \twolinecell{$-30^{\circ}$ - $+30^{\circ}$\\( $-135^{\circ}$ - $+135^{\circ}$)} \\
            {solarize}& \twolinecell{0 - 256\\(0 - 256)} & {{color}}& \twolinecell{0.1 - 1.9. \\ (0.01 - 2.)}\\
            posterize& \twolinecell{4 - 8\\(2 - 8)} & contrast& \twolinecell{0.1 - 1.9. \\ (0.01 - 2.)} \\
            {{brightness}}& \twolinecell{0.1 - 1.9. \\ (0.01 - 2.)} & {sharpness}& \twolinecell{0.1 - 1.9. \\ (0.01 - 2.)}\\
            shear\_x& \twolinecell{0.0 - 0.3\\0.0 - 0.99} & {shear\_y}& \twolinecell{0.0 - 0.3\\0.0 - 0.99}\\
            {{translate\_x}}& \twolinecell{0 - 10\\(0 - 32)} & {{translate\_y}}& \twolinecell{0 - 10\\(0 - 32)}\\
            \rowcolor{Added}
            \dashuline{cutout}& 0 - 0.2 & \dotuline{\dashuline{invert}}& -\\
            \rowcolor{Added}
            {flip\_lr}& - & flip\_ud& - \\
            \rowcolor{Added}
            sample\_pairing & 0.0 - 0.4 & blur & -\\
            \rowcolor{Added}
            smooth & - & &\\
    \end{tabular}
    }
    \caption{
    An overview of the augmentation spaces. The unmarked operations are shared by all augmentation spaces and make up the RA augmentation space. The \dashuline{UA} augmentation space additionally contains the dash underlined operations and the \dotuline{OHL} augmentation space additionally contains the dotted underlined operations. The ranges given here are the ones used for AA and RA with a discretization to thirty values. The UA augmentation space allows translation up to 14 pixels, but inherits all other settings from RA. The Wide augmentation space we use for batch augmentation has the same operations as RA, but uses the strength ranges in parantheses. The OHL augmentation space uses different ranges and a discretization to three values, see \cite{Lin2019ohlautoaug} for more details.
    All methods are defined as part of Pillow (\url{https://github.com/python-pillow/Pillow}), as part of ImageEnhance, ImageOps or as image attribute, besides cutout \cite{devries2017improved}. We also provide operations with the exact same names in our code.}
    \label{tab:searchspaces}
\end{table}

\section{Comparison of Different Methods on the Same Augmentation Space}
\label{sec:comwsameaugs}
While in the above experiments we used the augmentation space corresponding to each method in the evaluations, in this section, we probe the impact of these differences. We follow the setup of section \ref{section:experiments:reproducibility} and compare our reproduced results of each method to \ta{} on the exact same augmentation space as in the paper introducing the respective method. Table \ref{tab:compare_same_aug_space} shows that TA\textquotesingle{s} improvements generalize across augmentation spaces and methods.

\begin{table*}[h]
    \centering
    \setlength{\tabcolsep}{.3em}
    \begin{tabular}{ll|cccc}
         Dataset &Setup & AA & FAA & RA & UA  \\
         \hline
         CIFAR-10 & Method &  97.31 $\pm$ .22 & 97.43 $\pm$ .09 & 97.12 $\pm$ .14 & \textbf{97.46} $\pm$ .14\\ 
         & TA & \textbf{97.55} $\pm$ .06 & \textbf{97.51} $\pm$ .09 & \textbf{97.46} $\pm$ .09 & \textbf{97.42} $\pm$ .07\\
         \hline
         CIFAR-100 & Method &  82.91 $\pm$ .41  & \textbf{83.27} $\pm$ .13 & 83.1 $\pm$ .32 & 83.08 $\pm$ .27\\
         & TA & \textbf{83.34} $\pm$ .10 & \textbf{83.36} $\pm$ .15 &  \textbf{83.54} $\pm$ .12 & \textbf{83.33} $\pm$ .14\\
         \hline
         SVHN & Method & 97.99 $\pm$ .06 & - & \textbf{98.06} $\pm$ .04 & \textbf{98.05} $\pm$ .04 \\
         Core & TA & \textbf{98.04} $\pm$ .02 & 97.84 $\pm$ .03 & \textbf{98.05} $\pm$ .02 & \textbf{98.06} $\pm$ .04\\
    \end{tabular}
    \caption{Comparisons of various methods (in our reimplementation) to TA, using the exact same augmentation space.
    E.g., for CIFAR-10 on the AA space, AA reached 97.31 $\pm$ .22 and TA reached 97.55 $\pm$ .06. No policy is published for FAA on SVHN Core, since this setup was not part of the FAA paper. Therefore, we do not reproduce FAA on SVHN Core.}
    \label{tab:compare_same_aug_space}
\end{table*}


\section{Evaluation on Special Datasets}
\label{sec:specialdatasets}
\newcommand{\occc}{Occ.\ CIFAR-10}
To further show that this method generalizes to more particular image classification datasets without fine-tuning, we considered two more datasets, following the settings of Section \ref{section:experiments:reproducibility}. (i) Since we are not aware of an image recognition dataset that contains occlusions, we created an occlusion variant of CIFAR-10 (\occc{}), where a 14x14 square is occluded by a black box, in each image including the test images; we evaluate a WRN-28-10 on \occc{}. We follow the settings for CIFAR-10 closely for this experiment. (ii) Additionally, we evaluate an RN-50 on the Stanford Cars dataset, which is a dataset in which visual details are important to distinguish car models. We train for 1000 epochs. Table \ref{tab:specialdatasets} shows that TA continues to perform well in these settings, outperforming even brute-force tuned RA.

\begin{table}[h]
\footnotesize
    \setlength{\tabcolsep}{.3em}
    \centering
    \begin{tabular}{c|ccccc}
         Method & Baseline & Transfer-RA & BF-RA & TA (RA)  \\
         \hline
         \occc{} & 94.99 $\pm$ .11 & 95.2 $\pm$ .08 & 95.52 $\pm$ .18 & \textbf{95.72} $\pm$ .09 \\
         \hline
         Stanford Cars & 90.21 $\pm$ .16 & 92.47 $\pm$ .17 & - & \textbf{92.77} $\pm$ .12\\
         
    \end{tabular}
    \caption{A comparison on non-standard datasets. The RA setting of BruteForce-RA is searched in the same large set as in Section 4.1.2. Transfer-RA is transferred from Wide-ResNet-28-10 on CIFAR-10 and ResNet-50 on ImageNet, respectively. BF-RA for S.\ Cars was not feasible in the given time.}
    \label{tab:specialdatasets}
\end{table}

\section{Evaluation on EfficientNet-B1}
\label{sec:efficientneteval}
While we tried to evaluatate on as relevant setups as possible, we also had to make sure that we can compare with previous work for the main evaluation in Section \ref{section:comparison:fixedtrainingsetup}. Here, we add an Evaluation of an EfficientNet-B1 following the ImageNet setup described in the original paper \cite{tan2019efficientnet} closely.
None of the methods we compare to compares on this task, thus we re-implemented UA and RA as baselines. For RA we performed a search over 3 settings for $m$, namely 8, 14 and 21, and fixed the number of augmentations $n$ to 2 following the EfficientNet evaluations in \cite{cubuk2020randaugment}.

\begin{table}[]
\small
    \centering
    \begin{tabular}{c|cc|c}
         &  RA & UA & TA (Wide)\\
         \hline
           EfficientNet-B1 & 78.75 $\pm$ .16 & 78.83 $\pm$ .23 & 78.99 $\pm$ .12

    \end{tabular}
    \caption{The average performance of an EfficientNet-B1 across 5 re-runs on ImageNet with different augmentation methods.}
    \label{tab:my_label}
\end{table}

\section{Approximation of the Compute Costs for Different Methods}
\label{sec:computecostapproximation}
In this section, we discuss the data underlying our performance per compute comparison.
To fairly compare methods, we do not rely on published GPU times as much as possible, but instead calculate all costs for a RTX 2080 Ti for which we know many training times.
Therefore, we can only compare methods for which we ran the models.
That means for CIFAR-100 we consider only, the consider the Wide-ResNets as well as Shake-Shake-26-2x96d.

Our estimates for the cost of one epoch on the full CIFAR-10 dataset with 50,000 examples for each model:
\begin{itemize}
    \item Wide-ResNet-28-2: 16s
    \item Wide-ResNet-40-2: 40s
    \item Wide-ResNet-28-10: 101s
    \item Shake-Shake-26-2x96d: 83s
\end{itemize}

\paragraph{AA} In the AA paper \cite{cubuk2019autoaugment} the policy is trained over 15'000 evaluations of a Wide-ResNet-40-2 on 120 epochs of 4000 examples from CIFAR-10 for all models. 
We therefore estimate the search cost of AA as $15000\cdot 4000/50000\cdot 120\cdot40s = 1600h$.
Additionally we add the standard time for 200 epochs of standard training for each mode.
Wide-ResNet-40-2: $1600h+40s\cdot200/60/60$, Wide-ResNet-28-10: $1600h+101s\cdot 200/60/60$, Shake96: $1600h+83s\cdot 1800/60/60$.

\paragraph{Fast AA}
For Fast AA \cite{sungbin2019fastautoaugument}, we estimate, based on the GPU times in the paper that the search costs more than one full training. We therefore estimate the compute cost as one training.

\paragraph{UA and TA}
No search costs. Therefore the total cost simply is the cost of a single training. This is $\#\text{epochs}\cdot \text{costperepoch}$.

\paragraph{\advaa{} and TA x8}
We assume for both setups no costs, even though this of course is only a lower bound on the compute requirements of \advaa{}. We simply multiply the number of epochs with the cost per epoch and 8, the number of workers.

\paragraph{RA}
The authors of RA use a search space of 5 settings each is evaluated on 90\% of the full dataset with the same number of epochs and model. So we have a factor of $5\cdot 9/10$ with which we multiply the standard costs to get the search costs.

\paragraph{OHL}
OHL uses 300 epochs for the Wide-ResNets and trains with 8 parallel workers. We therefore have a factor of $8\cdot 300/200$ compared to standard costs for search and training combined.

\paragraph{AWS}
For AWS the data is not completely clear.
First, we have an earlier version of the paper that says it evaluates 800 policies, but in a later version this was corrected down to 500.
We therefore assume only 500 policy evaluations to be conservative. 
They used a Wide-ResNet-28-10 for the augmentation search CIFAR-100 experiments.
During augmentation search they train on 80\% of the training set for 200 epochs first, and then for 10 epochs for each policy evaluation. This yields $0.8\cdot (200+500\cdot 10)\cdot 101=117h$.

For AWS\textquotesingle{s} x8 setting (8-times augmented batches), we assume the same search costs as above and 8-times the training costs.

\begin{table*}[]
\small
    \centering
    \begin{tabular}{l|ccc}
         Method & Policies for Training & Code for training & Code for meta-training \\
         \hline
         \textbf{Cheap Search} \\
         TA & \cmark & \cmark & -\\
         UA & \cmark & \xmark & - \\
         Fast AA & \cmark$^\text{p}$ & \cmark & \cmark \\
         \textbf{Expensive Search ($>2\times$)} \\
         RA & \cmark & \xmark & \xmark \\
         Adv. AA & \xmark & \xmark & \xmark \\
         OHL & \xmark & \xmark & \xmark$^\text{p}$ \\
         \textbf{Very Expensive Search ($>10\times$)} \\
         PBA & \cmark$^\text{s}$ & \cmark$^\text{i}$ & \cmark$^\text{i}$ \\
         AWS & \xmark & \xmark & \xmark$^\text{p}$ \\
         AA & \cmark & \cmark$^\text{i}$ & \xmark  \\
    \end{tabular}
    \caption{In this table we compare the reproducibility of different methods in three categories. (i) Whether the augmentation policies used for  model trainings are available, (ii) whether the authors provide code for training a model with the policies on which they report their performance and (iii) whether there is code available to run the search for training policies, code for a meta-training. We mark entries with - if it is not an applicable category for the given augmentation method and additionally use the following symbols. \cmark$^\text{s}$: Only available for subset of experiments, \cmark$^\text{i}$: Not available for ImageNet trainings, which for PBA was also not considered in the paper, \xmark$^\text{p}$: there is publicly work in progress.}
    \label{tab:reproducibilityofpapers}
\end{table*}

\section{Recommendations for the Application of Automatic Augmentation Methods}
\label{sec:applicationrecommendation}
Based on our intense study of automatic augmentation methods for different image classification tasks using different models we recommend the following steps when applying automatic augmentation methods.
In the application of an automatic augmentation method it is of course important to know, whether a method is easy to reimplement. We thus put together Table \ref{tab:reproducibilityofpapers} for easy guidance.

\paragraph{Standard Model and Dataset}
If the model and dataset combination you are using is part of automatic augmentation literature, we recommend to simply use the best published method for your setup with published code and policies.

\paragraph{Novel model or Novel dataset}
If you are using a setup not evaluated in the automatic augmentation literature, it is a good approach to try both the best performing model on a similar task as well as a parameter-free baseline. The parameter-free baseline, like UniformAugment or TrivialAugment, especially can be expected to generalize to the new task, since they generalized to all standard automatic evluation benchmarks without any tuning.
If you have tuning budget, you can of course tune something like PBA to your particular task.
This likely is a good idea if your images are very dissimilar to the automatic augmentation benchmarks.

\end{document}